%% file: main.tex
\documentclass[10pt,twocolumn,letterpaper]{article}

\usepackage{iccv}              


\usepackage{tabularray}
\UseTblrLibrary{booktabs}
\usepackage{array}
\usepackage{multirow}
\usepackage[normalem]{ulem}
\usepackage{xcolor}     
\usepackage{colortbl}
\usepackage{makecell}
\usepackage{tabularx}

\definecolor{iccvblue}{rgb}{0.21,0.49,0.74}
\usepackage[pagebackref,breaklinks,colorlinks,allcolors=iccvblue]
{hyperref}

\def\paperID{8385} 
\def\confName{ICCV}
\def\confYear{2025}

\title{AMD: Adaptive Momentum and Decoupled Contrastive Learning Framework for Robust Long-Tail Trajectory Prediction}

\author{
    Bin Rao\textsuperscript{1}\thanks{Equal contribution.} \and
    Haicheng Liao\textsuperscript{1}\footnotemark[1] \and
    Yanchen Guan\textsuperscript{1} \and
    Chengyue Wang\textsuperscript{1} \and
    Bonan Wang\textsuperscript{1} \and
    Jiaxun Zhang\textsuperscript{1} \and
    Zhenning Li\textsuperscript{1}\thanks{Corresponding author: \texttt{zhenningli@um.edu.mo}} \and
    \textsuperscript{1}State Key Laboratory of Internet of Things for Smart City,
    University of Macau
}

\begin{document}
\maketitle
\vspace*{-10mm}
\input{sec/0_abstract}    
\input{sec/1_intro}

\input{sec/2_relatedwork}
\input{sec/3_methodology}

\input{sec/4_experiments}

\input{sec/5_conclusion}

\section*{Acknowledgements}
This work was supported by the Science and Technology Development Fund of Macau [0122/2024/RIB2 and 001/2024/SKL], the Research Services and Knowledge Transfer Office, University of Macau [SRG2023-00037-IOTSC, MYRG-GRG2024-00284-IOTSC], the Shenzhen-Hong Kong-Macau Science and Technology Program Category C [SGDX20230821095159012], the State Key Lab of Intelligent Transportation System [2024-B001], and the Jiangsu Provincial Science and Technology Program [BZ2024055].

{
    \small
    \bibliographystyle{ieeenat_fullname}
    \bibliography{main}
}

\end{document}

%% file: sec/0_abstract.tex
\begin{abstract}
Accurately predicting the future trajectories of traffic agents is essential in autonomous driving. However, due to the inherent imbalance in trajectory distributions, tail data in natural datasets often represents more complex and hazardous scenarios. Existing studies typically rely solely on a base model’s prediction error, without considering the diversity and uncertainty of long-tail trajectory patterns. We propose an adaptive momentum and decoupled contrastive learning framework (AMD), which integrates unsupervised and supervised contrastive learning strategies. By leveraging an improved momentum contrast learning (MoCo-DT) and decoupled contrastive learning (DCL) module, our framework enhances the model’s ability to recognize rare and complex trajectories. Additionally, we design four types of trajectory random augmentation methods and introduce an online iterative clustering strategy, allowing the model to dynamically update pseudo-labels and better adapt to the distributional shifts in long-tail data. We propose three different criteria to define long-tail trajectories and conduct extensive comparative experiments on the nuScenes and ETH/UCY datasets. The results show that AMD not only achieves optimal performance in long-tail trajectory prediction but also demonstrates outstanding overall prediction accuracy.
\end{abstract}

%% file: sec/1_intro.tex
\section{Introduction}
\label{sec:intro}

\begin{figure}[t]
  \centering
\includegraphics[width=0.45\textwidth]{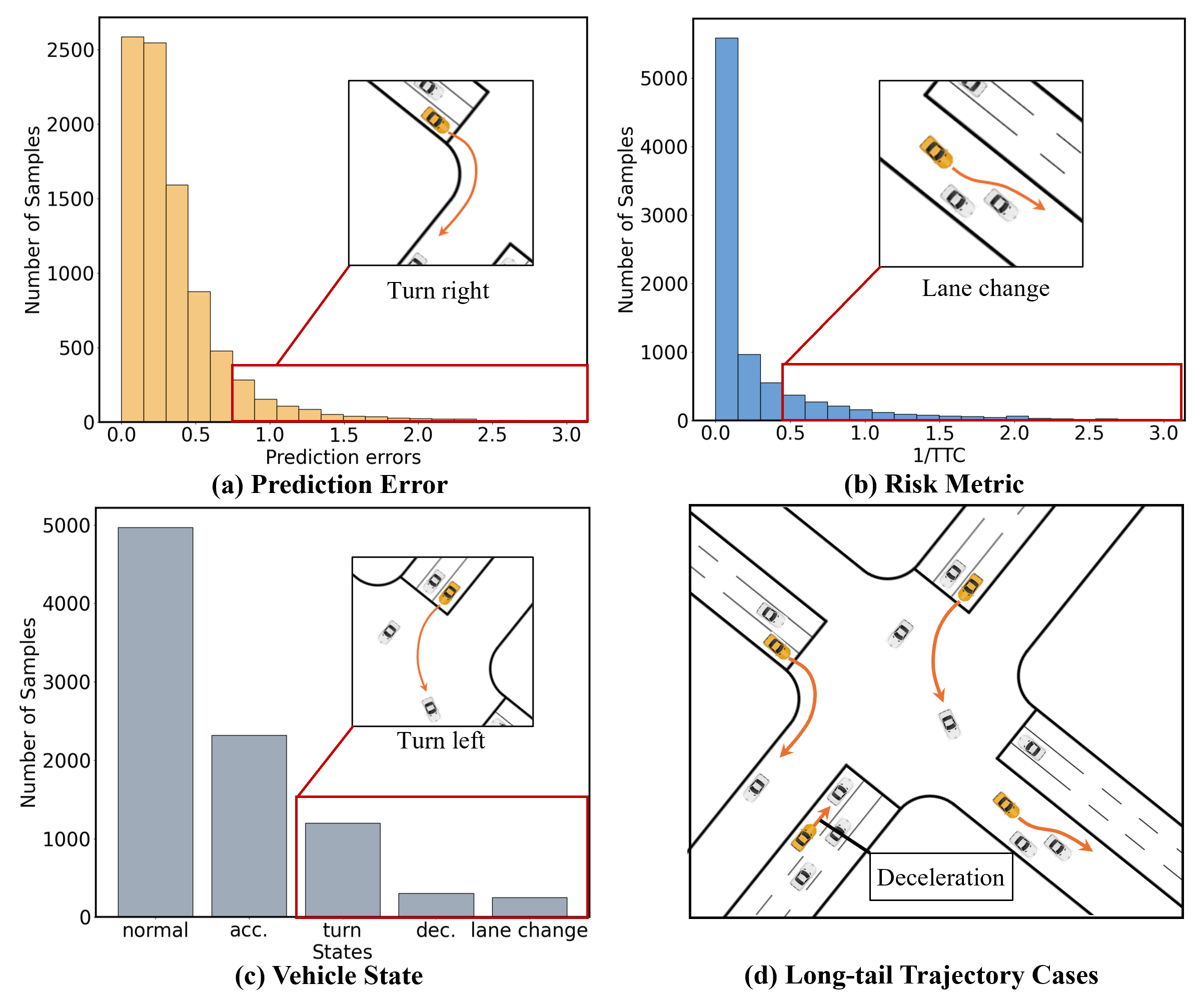} 
  \caption{Long-tail trajectory distributions defined from multiple perspectives. Panels (a), (b), and (c) illustrate distributions based on prediction error, the risk metric (inverse time-to-collision, 1/TTC), and vehicle state, respectively. Panel (d) presents vehicle trajectories under various scenarios—such as turning, lane changing, acceleration (acc.), and deceleration (dec.) maneuvers—offering a visual representation to facilitate understanding of long-tail trajectories.}
  \label{fig1} 
\vspace{-6pt}
\end{figure}

Achieving high-level autonomous driving relies heavily on the ability to accurately predict the future trajectories of surrounding traffic agents \cite{liao2025cot, WOS:001270539300037, wang2025dynamics}. Precise trajectory prediction enables autonomous vehicles to make informed decisions, ensuring safety and efficiency in complex traffic environments. Despite significant advancements, autonomous systems still face challenges in handling the vast diversity of driving behaviors exhibited in real-world scenarios.

One critical challenge arises from the \textbf{long-tail distribution} of driving behaviors. In real-world traffic, a small number of common behaviors—such as steady cruising or standard lane changes—occur frequently and dominate datasets. In contrast, numerous rare but potentially hazardous behaviors, like abrupt maneuvers or interactions with erratic agents, are underrepresented. This imbalance poses significant difficulties for trajectory prediction models, which tend to perform well on frequent behaviors but struggle with rare ones. Addressing this issue is essential for the safe and reliable operation of autonomous vehicles.

To advance the field, it is crucial to address several fundamental questions that have been inadequately explored:

\textbf{Q1: What exactly constitutes the long-tail challenge in trajectory prediction?} While data imbalance is a recognized issue, prior studies often lack a precise definition of the long-tail challenge in the context of trajectory prediction for autonomous driving. The long-tail phenomenon refers to a statistical distribution where a few common events (the ``head") comprise the majority of data, while many rare events (the ``tail") have few instances \cite{reed2001pareto, li2022longtaildis}. In trajectory prediction, this means datasets are rich in common driving behaviors but sparse in rare ones. These rare behaviors, however, are often critical for safety, involving complex maneuvers or high-risk situations \cite{li2023graph-risk, thuremella2024risk,liao2025toward}.

\textbf{Q2: How can we effectively identify and characterize long-tail trajectories within datasets?} Identifying which trajectories constitute the long tail is challenging due to the lack of explicit labels and the diversity of rare behaviors. Existing methods often rely on model-specific prediction errors to infer long-tail instances \cite{zhou2022long, lan2024hi-scl}, which can be inconsistent across models and insufficient for capturing rare behaviors. A systematic approach is needed to identify long-tail trajectories based on intrinsic properties, such as risk levels, maneuver complexity, or other meaningful criteria. This would enable a more comprehensive understanding of underrepresented behaviors that are critical to safety. 

\textbf{Q3: How can we design models that accurately predict long-tail trajectories without compromising overall performance?} Efforts to improve prediction accuracy on rare trajectories often face a trade-off with performance on common behaviors \cite{salzmann2020trajectron++, liao2024bat-trans}. Focusing excessively on the tail can lead to overfitting or neglect of the head, reducing overall model effectiveness. Therefore, it is imperative to develop learning strategies that can enhance the prediction of rare trajectories while maintaining or even improving performance on common ones. 

To address these critical questions, we propose a comprehensive approach to systematically tackle the long-tail challenge in trajectory prediction. We begin by formally defining the long-tail distribution in the context of autonomous driving by analyzing real-world driving data, illustrating how the imbalance manifests, and discussing its implications for model performance and safety. Next, we introduce a multi-criteria method for identifying long-tail trajectories based on intrinsic properties such as prediction error distribution, risk metrics like low time-to-collision (TTC), and complex vehicle states. By integrating these perspectives, as shown in Figure \ref{fig1}, we comprehensively characterize long-tail trajectories, providing a holistic foundation for developing targeted strategies to improve prediction accuracy in these challenging cases. To address the modeling challenges, we propose AMD, an \textbf{A}daptive \textbf{M}omentum and \textbf{D}ecoupled Contrastive Learning Framework designed to enhance prediction accuracy on long-tail trajectories without compromising overall performance. AMD incorporates adaptive momentum updating to emphasize underrepresented samples, decoupled contrastive learning to balance optimization between head and tail classes, innovative data augmentation strategies to simulate real-world uncertainties and an online iterative clustering mechanism to adapt to distributional changes in the data.


Our work makes the following contributions:

1) We develop a multi-criteria method to identify and characterize long-tail trajectories based on intrinsic properties, enabling targeted improvements in prediction models. Defining long-tail trajectories by prediction error, risk metrics, and vehicle states ensures the model can effectively handle diverse and complex scenarios.

2) We propose an adaptive and robust framework that effectively balances learning between common and rare trajectories, enhancing prediction accuracy on long-tail data without degrading overall performance.

3) We conduct comprehensive experiments that consistently demonstrate AMD's superior performance in terms of accuracy, adaptability, and reliability compared to existing state-of-the-art (SOTA) methods, validating its effectiveness across various challenging scenarios.

%% file: sec/2_relatedwork.tex
\section{Related Work}
\label{sec:Related Work}

\begin{figure*}[t]
  \centering
\includegraphics[width=1\textwidth]{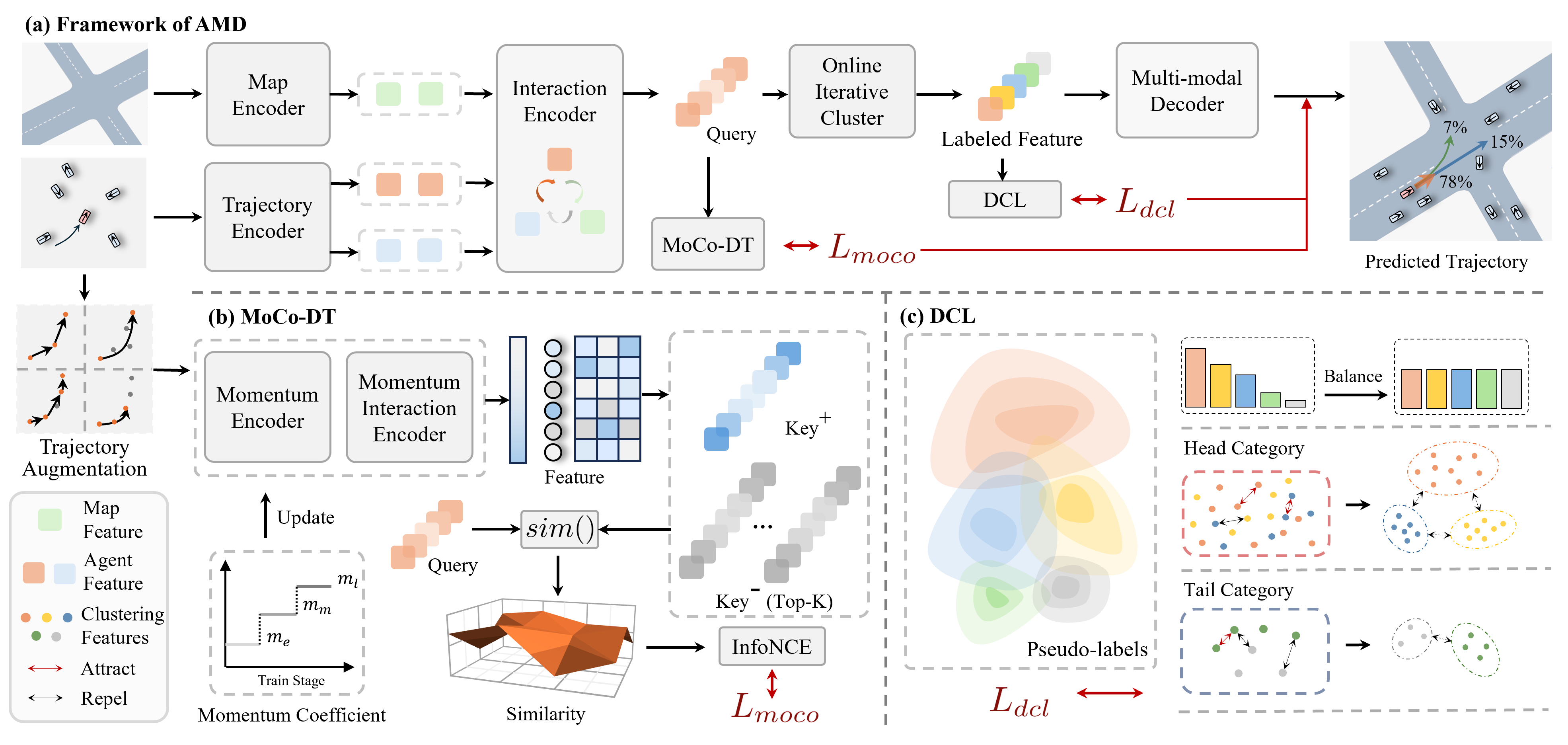} 

  \caption{Overview of the proposed AMD framework. Panel (a) illustrates the structure of the model, including the Encoder, Interaction Module, and Predictor, which collectively enable multimodal trajectory prediction. This model takes as input the target agent, surrounding agents, and HD maps, ultimately outputting predicted multimodal trajectories. Panels (b) and (c) present details of the Adaptive Momentum Contrastive Learning (MoCo-DT) module design and the Decoupled Contrastive Learning (DCL) module.}
  \label{fig2} 
\end{figure*}

Trajectory prediction is a key challenge in autonomous driving. Early methods based on kinematic and statistical models \cite{lin2000vehicle-yuce-1,wong2022view-yuce-2} are computationally efficient but struggle with complex environmental influences, limiting accuracy. Driven by data-centric approaches such as VectorNet \cite{gao2020vectornet}, deep learning models have shown remarkable potential in trajectory prediction. Architectures such as Recurrent Neural Networks (RNNs) \cite{alahi2016social-lstm,huang2021bayonet-GRU,liao2025minds,wang2025beyond}, Graph Neural Networks (GNNs) \cite{xu2022adaptive-GNN,liao2024cognitive,liao2024mftraj,wang2025nest,wang2025wake}, and Transformers \cite{liao2024bat-trans,salzmann2020trajectron++,mohamed2020social,liao2024cdstraj} have significant strengths in modeling temporal and spatial dependencies. However, capturing the inherent uncertainty in vehicle motion remains a major challenge in trajectory prediction.

\subsection{Long-Tail Trajectory Prediction}

Although existing trajectory prediction models perform well on benchmark datasets, they often struggle with rare or challenging scenarios—an issue known as the long-tail challenge in trajectory prediction \cite{zhou2022long,liao2024physics}. In data-driven deep learning models, prediction performance heavily relies on data quality, and the inherent data imbalance exacerbates the long-tail problem \cite{li2022longtaildis}. This issue is not unique to trajectory prediction and is observed across various domains, such as image classification and natural language processing \cite{longtail-1,longtail-2}. Numerous strategies, including data resampling \cite{han2005borderline-resample} and loss re-weighting \cite{ross2017focal-loss}, have been proposed to address this problem. Recently, some studies have specifically targeted long-tail trajectory prediction. For example, FEND \cite{wang2023fend} framework enhances long-tail prediction by augmenting future trajectories. TrACT \cite{zhang2024tract} architecture identifies long-tail trajectories based on training curves. Hi-SCL \cite{lan2024hi-scl} represents traffic scenarios as waveforms to improve feature extraction in long-tail trajectory prediction. These efforts underscore the increasing importance and interest in developing specialized techniques to effectively address the long-tail problem in trajectory prediction.


\subsection{Contrastive Learning}

Traditional contrastive learning \cite{chen2020simple-clr} is an unsupervised strategy that compares different views of data to learn similar and distinct features, forming effective representations. Momentum Contrast (MoCo) \cite{he2020momentum-moco} is an unsupervised contrastive learning framework that overcomes the issue of updating the contrastive sample pool by maintaining a dynamic memory, and enhancing representation learning.

Supervised contrastive learning \cite{khosla2020supervised-cl,xuan2024decoupled} adds label information to guide the construction of positive and negative pairs, improving the model's ability to group similar samples and distinguish dissimilar ones. This method is more robust than unsupervised learning. In trajectory prediction, various frameworks have applied contrastive learning to enhance learning on underrepresented samples \cite{chang2023contrastive,lan2024hi-scl,wang2023fend}. However, these methods often overlook the specific challenges of recognizing and predicting long-tail trajectories. Unlike previous methods, our study proposes a dual-layer contrastive framework combining unsupervised and supervised strategies, enhancing long-tail trajectory recognition and enabling more accurate predictions.

%% file: sec/3_methodology.tex
\section{Methodology}
\label{sec:Methodology}
\textbf{Problem Formulation}. Trajectory prediction is a typical temporal sequence prediction problem. Given a traffic scenario containing \( n+1 \) agents, the motion of all agents is represented as a series of state sequences, denoted as \(\{ X_i, Y_i\} \) (\( i \in [0, n] \)). Here, \( X_i = \{ X_i^t | t \in [0, T_h] \} \) represents the past observed trajectory of agent \( i \), which includes information such as the agent's position, speed, and heading angle of agent \( i \) and its surrounding agents, and \( Y_i = \{ Y_i^t | t \in [T_h, T_h+T_f] \} \) represents the future trajectory of agent \( i \), also known as the ground truth. The road map information is represented as a series of feature vectors \( M_N = \{ m_1,...,m_N \}\). Therefore, the trajectory prediction task is defined as predicting the future trajectory \( Y_i \) of the target agent based on the observed past trajectory \( X_i \) and environmental context \( M_N \).\\
\textbf{Overview}. The overall framework of AMD described in Figure \ref{fig2}, starts with four augmentation strategies that randomly enhance the target agent’s trajectory. Both the original and augmented trajectories are processed by a feature encoder, producing high-dimensional feature representations. A scene interaction module employing self-attention and cross-attention mechanisms integrates the target agent's features with those of surrounding agents and maps data to create a contextualized representation. These features are then used as positive samples in an improved momentum contrastive learning module (MoCo-DT). An online iterative clustering strategy generates pseudo-labels, and input into a decoupled contrastive learning (DCL) module. Finally, a multi-modal trajectory decoder predicts future trajectories across different modes.

\subsection{Trajectory Augmentation}
To address data imbalance and enhance model generalization, we draw inspiration from \cite{chang2023contrastive} and propose four novel augmentation methods for short-term trajectories: (1) Simplify, (2) Shift, (3) Mask, and (4) Subset. These methods are designed to improve the model’s robustness against real-world uncertainties in trajectory prediction and enhance accuracy for long-tail trajectories. Specifically: (1) Simplify reduces redundant points to emphasize primary movement patterns; (2) Shift applies random displacements to simulate external perturbations; (3) Mask randomly discards data points to mimic sensor failures; and (4) Subset selects consecutive subsequences to emulate incomplete temporal information. By generating diverse trajectory transformations, these methods enable the model to better adapt to varying patterns during training. Our experiments further confirm that such targeted augmentations lead to substantial performance gains in challenging scenarios. Detailed implementations are provided in \textbf{Appendix}.

\subsection{Trajectory Feature Extractor}
\textbf{Feature Encoder}. The trajectory encoder converts raw and augmented trajectory data into high-dimensional representations, providing quality feature inputs for subsequent tasks. We use a hierarchical embedding method combining Multi-Layer Perceptron (MLP), a Transformer Encoder (TrEnc), and Gated Recurrent Unit (GRU) to capture trajectory temporal features. To model driver memory decay, the final GRU hidden states are used as encoding features for the target agent $F_{{tar}}$ and neighboring agents $F_{{nbr}}$. For the HD map $M$, lane nodes and lines are represented as discrete vectors \cite{gao2020vectornet}, and hierarchical extraction yields the high-dimensional map encoding $F_{{lane}}$.
\begin{equation}
F_{\text{tar}} = \phi_{G}\left(\phi_{T}\left(\phi_{M}(t_{\text{tar}})\right) + \phi_{M}(t_{\text{tar}})\right)
\end{equation}
\begin{equation}
F_{\text{nbr}} = \phi_{G}\left(\phi_{T}\left(\phi_{M}(t_{\text{veh}}, t_{\text{ped}})\right) + \phi_{M}(t_{\text{veh}}, t_{\text{ped}})\right)
\end{equation}
\begin{equation}
F_{\text{lane}} = \phi_{A}\left(\phi_{G}\left(\phi_{M}(l_{\text{node}})\right), A\right)
\end{equation}
where $\phi_{M}$, $\phi_{T}$, $\phi_{G}$, and $\phi_{A}$ denote embedding functions implemented by the MLP, TrEnc, GRU and GAT. $t_{{tar}}$, $t_{{veh}}$, $t_{{ped}}$, and $l_{{node}}$ represent the input features of the target agent, surrounding vehicles, pedestrians, and lane nodes, respectively. $A$ is the adjacency matrix of the lane nodes.\\
\textbf{Scene Interaction Encoder}. To capture the dynamic interactions between the target agent and its surrounding environment, we designed a scene interaction encoder that incorporates a latent variable mechanism to generate diverse potential trajectory patterns. This module uses a unified cross-modal attention mechanism to fuse information from multiple modalities, including target features, surrounding agent features, and road node features. The result is a set of multimodal fused features $F_{cross}$, generated through positional encoding($pos$) and multi-head attention ($MHA$).
\begin{equation}
F_{cross}=MHA(H_{mode}+pos,K_{t+n+l},V_{t+n+l})
\end{equation}
where $H_{mode}$ represents the pattern features generated by the latent variable mechanism, and $K_{t+n+l}$ and $V_{t+n+l}$ are the merged key and value, which include features of the target, surrounding agents, and road nodes. Further details are in the \textbf{Appendix}.

\subsection{Momentum Contrastive Learning}

In long-tail prediction tasks, rare samples are often overlooked due to their scarcity. To address this, we propose an improved Dynamic Momentum and Top-K Hard Negative Mining method (MoCo-DT), which improves the focus on long-tail samples in contrastive learning. Compared to the original MoCo approach \cite{he2020momentum-moco}, MoCo-DT dynamically adjusts the momentum coefficient $m$ based on training progress $t$ and total duration $T$, employing distinct coefficients $m_e, m_m, m_l$ for the early, middle, and late training stages, respectively, to adapt to different training stages.

The Top-K Hard Negative Mining mechanism strengthens the model’s ability to distinguish challenging long-tail samples by selecting the most similar negative samples. Specifically, this mechanism computes the similarity between the query sample and both positive and negative samples, dynamically selecting the Top-K hardest negatives from the negative set to emphasize learning from difficult long-tail features in contrastive training.
\begin{equation}
\begin{aligned}
& l_{{pos}} = {sim}(q, k^+), \quad l_{{neg}} = {sim}(q, k^-) \\
& \{k_1^-, \dots, k_K^-\} = {TopK}_{k^-}({sim}(q, k^-))
\end{aligned}
\end{equation}
where \(sim()\) denotes the similarity function, \( q \) is the query encoding, and \( k^+ \) and \( k^- \) represent positive and negative encodings, respectively. \( K \) is the number of hard negative samples selected, and \({TopK}\) refers to selecting the top \( K \) samples with the highest similarity to the query sample among the negative samples.

The final contrastive loss is constructed by comparing the positive sample with the Top-K hard negative samples, calculated as follows:
\begin{scriptsize}
\begin{equation}
L_{{moco}} = -\log \frac{\exp({sim}(q, k^+)/\tau)}{\exp({sim}(q, k^+)/\tau) + \sum_{i=1}^K \exp({sim}(q, k_i^-)/\tau)}
\end{equation}
\end{scriptsize}

\noindent where \( \tau \) is a temperature parameter. This design allows the model to focus on challenging long-tail features.

\subsection{Online Iterative Clustering Strategy}

Traditional methods typically use static clustering on encoder-derived features with fixed labels, which struggle to adapt to dynamic feature distributions, especially in long-tail data. To address this, we propose an Online Iterative Clustering Strategy that dynamically updates pseudo-labels during training to improve recognition of long-tail patterns.

This strategy involves clustering sample features in each training epoch to generate adaptive pseudo-labels. After each mini-batch, target feature representations are stored in a feature set. At predefined intervals, K-means clustering \cite{hartigan1979algorithm-kmeans} is applied to this set to create clusters representing distinct trajectory patterns, with clustering outcomes serving as pseudo-labels. By continuously updating these labels, our approach effectively captures subtle variations in rare trajectories, improving robustness and accuracy in long-tail trajectory identification.

\subsection{Decoupled Contrastive Learning}
Decoupled Contrastive Learning (DCL) \cite{xuan2024decoupled} is a type of supervised contrastive learning approach. Compared with traditional contrastive methods, DCL mitigates the bias toward head classes with high frequency, thus enhancing the prediction performance on long-tail data. By assigning different weights to positive samples from two categories, DCL achieves a balanced representation of both head and tail classes. DCL employs an L2 regularization to prevent the optimization from being influenced by class sample sizes. This approach effectively maximizes inter-class distance while minimizing intra-class distance, ensuring robust performance across head and tail classes. The DCL loss function is defined as:
\begin{small}
\begin{equation}
L_{dcl} = \frac{-1}{|P_i| + 1} \sum_{q_t \in \{q_i^+, P_i\}} \log \frac{\exp(w_r \cdot \langle q_t, q_i \rangle / \tau)}{\sum_{q_m \in \{q_i^+, U_i\}} \exp(\langle q_i, q_m \rangle / \tau)}
\end{equation}
\end{small}
where \( q_i \) and \( q_i^+ \) are features of positive samples in the same category, and \( U_i \) is the set of all other category features. \( P_i \) denotes the set of features in a given category, \( \tau \) is a temperature parameter, and \( w_r \) is the weight defined as:
\begin{equation}
w_r = 
\begin{cases} 
\alpha(|P_i| + 1), & \text{if } q_i = q_i^+ \\ 
(1 - \alpha)(|P_i| + 1) / |P_i|, & \text{if } q_i \in P_i 
\end{cases}
\end{equation}
\noindent where \( \alpha \in [0,1] \) is a hyperparameter balancing the weighting between within-category and inter-category samples.

\subsection{Multi-modal Decoder}
In long-tail trajectory prediction, the target agent may have multiple possible future paths. To address this, we design a Multi-modal Decoder that captures diverse trajectory modes using a latent variable mechanism and a Laplace Mixture Density Network (Laplace MDN). A single-layer GRU decoder generates varied trajectories, with the Laplace MDN outputting position, scale parameters, and probabilities to assess each mode’s likelihood.

\subsection{Training Loss}
For multi-modal trajectory prediction, we employ a Laplace negative log-likelihood as the regression loss \( L_{{reg}} \) and a cross-entropy loss \( L_{{cls}} \) for mode classification, with direct training loss $L_{{target}}$ as the task loss \( L_{{task}} \). To further enable the model to capture long-tail trajectory characteristics, we incorporate momentum contrastive loss and decoupled contrastive loss to ensure the accuracy of trajectory prediction. The final total loss $L$ is defined as follows:
\begin{equation}
L_{{task}} = L_{{target}} + \gamma_1 L_{{reg}} + \gamma_2 L_{{cls}}
\end{equation}
\begin{equation}
L = L_{{task}} + \lambda_1 L_{{moco}} + \lambda_2 L_{{dcl}}
\end{equation}
where \( \gamma_1 \), \( \gamma_2 \), \( \lambda_1 \), and \( \lambda_2 \) are weighting parameters.

%% file: sec/4_experiments.tex
\section{Experiments}
\label{sec:Experiments}

\begin{table*}[htbp]
\centering
\footnotesize
\begin{tabular}{lcccccccc} 
\toprule
Dataset & Model                   & Top 1\%                   & Top 2\%                   & Top 3\%                   & Top 4\%                   & Top 5\%                   & Rest                      & All                         \\ 
\midrule
\multirow{5}{*}{nuScenes} &Traj++ EWTA \cite{makansi2021on-exposing}            & 1.73/4.43                & 1.36/3.54                & 1.17/3.03                & 1.04/2.68                & 0.95/2.41                & 0.16/0.26                 & 0.22/0.39                   \\ 

& Traj++ EWTA+contrastive \cite{makansi2021on-exposing} & 1.28/2.85                 & 0.97/2.15                 & 0.83/1.83                 & 0.76/1.64                 & 0.70/1.48                 & \uline{0.15}/0.24                 & \uline{0.18}/0.30           \\ 

& FEND \cite{wang2023fend}                   & \uline{1.21}/\uline{2.50} & \uline{0.92}/\uline{1.88} & \uline{0.79}/\uline{1.61} & \uline{0.72}/\uline{1.43} & \uline{0.66}/\uline{1.31} & \textbf{0.14}/\uline{0.20} & \textbf{0.17}/\uline{0.26}  \\ 

& TrACT \cite{zhang2024tract}                  & 1.23/2.65                & 0.98/2.11                & 0.85/1.82                & 0.78/1.64                & 0.72/1.49                & -                         & 0.19/0.31                  \\ 

\rowcolor{green!10} & AMD (ours)              & \textbf{1.08}/\textbf{1.66 }       & \textbf{0.85}/\textbf{1.33 }       & \textbf{0.75}/\textbf{1.15 }       & \textbf{0.69}/\textbf{1.03 }       & \textbf{0.64}/\textbf{0.95}        & 0.18/\textbf{0.16}        & 0.21/\textbf{0.21}          \\
\midrule
\multirow{8}{*}{ETH/UCY}  & Traj++ EWTA \cite{makansi2021on-exposing}            & 0.98/2.54                & 0.79/2.07                & 0.71/1.81                & 0.65/1.63                & 0.60/1.50                &\textbf{0.14}/\uline{0.26}                 & 0.17/0.32                   \\ 
         & Traj++ EWTA+resample \cite{shen2016relay} & 0.90/2.17                & 0.77/1.90                & 0.73/1.78                & 0.66/1.60                & 0.64/1.52                & 0.20/0.41                 & 0.23/0.47                   \\
         & Traj++ EWTA+reweighting \cite{cui2019class} & 0.97/2.47 & 0.78/2.03 & 0.68/1.73 & 0.62/1.55 & 0.56/1.40 & 0.15/0.26 & 0.18/0.32 \\
         
         & Traj++ EWTA+contrastive \cite{makansi2021on-exposing} & 0.92/2.33 & 0.74/1.91 & 0.67/1.71 & 0.60/1.48 & 0.55/1.32 & 0.15/0.27 & 0.17/0.32 \\
         & LDAM \cite{cao2019learning}                 & 0.92/2.35                & 0.76/1.96                & 0.68/1.71                & 0.62/1.53                & 0.57/1.37                & 0.15/0.27                 & 0.17/0.33                   \\ 
         & FEND \cite{wang2023fend}                   & 0.84/2.13                & 0.68/1.68                & 0.61/1.46                & 0.56/\uline{1.30}                & 0.52/1.19                & \uline{0.15}/0.27                 & \uline{0.17}/0.32                   \\ 
                          & TrACT \cite{zhang2024tract}                  & \uline{0.80}/\uline{2.00} & \textbf{0.65}/\uline{1.63} & \uline{0.61}/\uline{1.46} & \uline{0.56}/1.31 & \uline{0.52}/\uline{1.18} & \uline{-}                  &\textbf{0.17}/\uline{0.32}   \\ 
         \rowcolor{green!10} & AMD (ours)              & \textbf{0.76}/\textbf{1.75}                 & \uline{0.66}/\textbf{1.59}                        & \textbf{0.58}/\textbf{1.37}                        & \textbf{0.54}/\textbf{1.25}                       & \textbf{0.51/1.16}     & 0.16/\textbf{0.24}                   & 0.18/\textbf{0.27}                            \\
\bottomrule
\end{tabular}
\caption{
Prediction errors (minADE/minFDE) for seven test samples from the nuScenes and ETH/UCY datasets, categorized by prediction error (FDE). For comparison with other methods, the nuScenes dataset uses a prediction horizon of 2s, while the ETH/UCY dataset uses a prediction horizon of 4.8s. The Top 1\%-5\% refers to the subset of samples with the largest prediction errors. Bold and underlined text represent the best and second-best results, respectively. Cases marked with ('-') indicate missing values.}
\label{table_1}%
\vspace{-6pt}
\end{table*}

\subsection{Experimental Setup}
\textbf{Datasets}. We evaluate our proposed method on the nuScenes \cite{caesar2020nuscenes} and ETH/UCY \cite{pellegrini2009you, leal2014learning} datasets, which contain real-world traffic data for vehicle and pedestrian scenarios, respectively, covering diverse trajectory patterns.

\noindent \textbf{Long-tail Subset}. To validate our model on long-tail data, we divide the dataset using three distinct criteria, differing from previous studies in long-tail trajectory prediction:
\begin{itemize}
    \setlength{\leftskip}{0.5em}
     \item \textbf{Prediction Error}: The dataset is divided into seven subsets based on prediction error: the Top 1\%-5\% with the highest errors, the remaining samples, and all samples.
    
    \item \textbf{Risk Metric}: We use (TTC) as a risk metric, identifying the Top 1\%-3\% of samples with the lowest TTC values, representing high-risk scenarios for target agents.
    
    \item \textbf{Vehicle State}:  We categorize samples based on the target agent’s behavior, specifically labeling rapid acceleration, rapid deceleration, sharp lane changes, and sharp turns, creating four distinct long-tail subsets.
\end{itemize}

This multi-criteria approach avoids the limitations of single criteria definitions for long-tail trajectories and provides a more comprehensive evaluation of the model's performance on diverse long-tail trajectories.

\noindent \textbf{Metrics}. We evaluate trajectory prediction performance using Average Displacement Error (ADE), Final Displacement Error (FDE), and Miss Rate (MR). For long-tail samples, we use minimum ADE (minADE) and minimum FDE (minFDE) to better assess performance on challenging samples. For overall multi-modal prediction, we use $\text{minADE}_k$ and $\text{minFDE}_k$ to evaluate the top-K predicted trajectories. 

\noindent \textbf{Implementation Details}. In our experiments, the loss function weight parameters are set to $\gamma_1 = 1$, $\gamma_2 = 0.5$, $\lambda_1 = 1$, and $\lambda_2 = 0.1$. For MoCo-DT, the parameters $m_e$, $m_m$, and $m_l$ are set to 0.95, 0.99, and 0.999, respectively. All models are trained on an NVIDIA RTX 3090 GPU. For additional experimental setup details, please refer to the \textbf{Appendix}.

\subsection{Comparisons to SOTA} 
\indent \textbf{(i) Quantitative Comparison under Prediction Error}.
To demonstrate the effectiveness of our method, we compared it with state-of-the-art long-tail trajectory prediction models. As shown in Table \ref{table_1}, our method outperforms other models on the Top 1\%-5\% most challenging long-tail samples. For the Top 1\% hardest samples, our method achieves an error of 1.08/1.66, reducing error by 14.9\% and 33.6\% compared to the closest competing model. Additionally, for all samples (All), our method also demonstrates superior performance, reducing the minFDE by 19.2\% compared to the closest competing model. These results highlight AMD’s advantages in long-tail trajectory prediction and its competitive edge in overall accuracy, with high precision and consistency across different levels of sample difficulty.

\noindent \textbf{(ii) Quantitative Results under Risk Metric and Vehicle State}.
We further validated our method’s performance on long-tail scenarios using Risk Metric and Vehicle State subsets from the nuScenes dataset (Tables \ref{table_2} and \ref{table_3}). Table \ref{table_2} shows that the prediction error for the top 1\%-3\% high-risk subset slightly exceeds that of the overall sample, suggesting effective adaptation to high-risk trajectories. In Table \ref{table_3}, four long-tail subsets involving emergency maneuvers exhibit higher prediction errors compared to normal trajectories, underscoring the difficulty of predicting sharp lane changes and turns. Despite these challenges, our model consistently demonstrates robust predictive accuracy across diverse long-tail conditions.

\begin{table}[!htbp]
\centering
\footnotesize
\begin{tabular}{ccccc} 
\toprule
\makecell{Prediction \\ Horizon (s)} & Top 1\%   & Top 2\%   & Top 3\%   & All        \\
\midrule
2                      & 0.20/0.20 & 0.24/0.24 & 0.24/0.23 & 0.21/0.21  \\
4                      & 0.40/0.44 & 0.44/0.48 & 0.43/0.48 & 0.42/0.48  \\
6                      & 0.65/0.72 & 0.72/0.83 & 0.70/0.87 & 0.69/0.88  \\
\bottomrule
\end{tabular}
\vspace{-6pt}
\caption{Prediction errors (minADE/minFDE) at different prediction horizons on nuScenes dataset, categorized by risk level. The top 1\%-3\% are the highest-risk subsets of the test samples.}
\label{table_2}%
\vspace{-6pt}
\end{table}

\begin{table}[htbp]
\centering
\scriptsize
\begin{tblr}{
    column{2-7} = {wd=0.7cm}, 
    cells = {c},
    rowsep = 1pt,
}
\toprule
\makecell{Prediction \\ Horizon (s)} & RA & RD & SLC & ST & Normal    & All        \\
\midrule
2                      & 0.29/0.27          & 0.34/0.31          & 0.39/0.40         & 0.32/0.34  & 0.19/0.18 & 0.21/0.21  \\
4                      & 0.51/0.57          & 0.57/0.63          & 0.68/0.74         & 0.58/0.71  & 0.36/0.42 & 0.42/0.49  \\
6                      & 0.80/1.01          & 0.90/1.08          & 1.10/1.41         & 0.94/1.28  & 0.61/0.78 & 0.69/0.88  \\
\bottomrule
\end{tblr}
\vspace{-6pt}
\caption{Prediction errors (minADE/minFDE) at different prediction horizons on nuScenes dataset, categorized by vehicle state for the six test samples. RA: Rapid Acceleration, RD: Rapid Deceleration, SLC: Sharp Lane Change, ST: Sharp Turn.}
\label{table_3}%
\vspace{-6pt}
\end{table}

\begin{table}[!htbp]
\centering
\footnotesize
\begin{tabular}{ccccc}
\toprule
Model         & minADE${_5}$       & $\text{minADE}_{10}$      & minFDE$_1$       & MR$_5$            \\
\midrule
Trajectron++ \cite{salzmann2020trajectron++}         & 1.88          & 1.51          & 9.52          & 0.70           \\
P2T \cite{deo2020trajectory-P2T}                  & 1.45          & 1.16          & -             & 0.64           \\
LaPred \cite{kim2021lapred}                    & 1.47          & 1.12          & 8.12          & \uline{0.53}   \\
GoHome \cite{gilles2022gohome}                 & 1.42          & 1.15          & \uline{6.99} & 0.57           \\
ContextVAE \cite{xu2023context-VAE} & 1.59          & -             & 8.24          & -              \\
SeFlow \cite{zhang2025seflow}                 & \uline{1.38}  & \textbf{0.98} & 7.89          & 0.60           \\
AFormer-FLN \cite{xu2024adapting-AFormer}                & 1.83          & 1.32          & -             & -              \\
\rowcolor{green!10} AMD(ours)                                  & \textbf{1.23} & \uline{1.06}  & \textbf{6.99}  & \textbf{0.50}  \\
\bottomrule
\end{tabular}
\vspace{-6pt}
\caption{Comparison of the performance of various models across all samples on nuScenes dataset, using 6s trajectory predictions.}
\label{table_4}%
\vspace{-12pt}
\end{table}

\begin{table*}[htbp]
\centering
\footnotesize
\begin{tabularx}{0.85\textwidth}{c *{8}{>{\centering\arraybackslash}X}}
\toprule
Model          & Venue           & ETH                     & HOTEL                   & UNIV                    & ZARA1                   & ZARA2                   & AVG                      \\ 
\midrule
PECNet \cite{mangalam2020not}         & ECCV           & 0.54/0.87               & 0.18/0.24               & 0.22/0.39               & 0.17/0.30               & 0.35/0.60               & 0.29/0.48                \\ 
AgentFormer \cite{yuan2021agentformer}     & ICCV           & 0.45/0.75               & 0.14/0.22               & 0.25/0.45               & 0.18/0.30               & 0.14/0.24               & 0.23/0.39                \\ 
Trajectron++ \cite{salzmann2020trajectron++}   & ECCV          & 0.39/0.83               & 0.12/0.21               & \uline{0.20}/0.44               & \textbf{0.15}/0.33               & \textbf{0.11}/0.25               & \uline{0.19}/0.41                \\ 

NPSN \cite{bae2022non}       & CVPR              & 0.36/0.59 & 0.16/0.25               & 0.23/\uline{0.39}               & 0.18/0.32               & 0.14/0.25               & 0.21/0.36                \\ 
MID \cite{gu2022stochastic}       & CVPR               & 0.39/0.66               & 0.13/0.22 & 0.22/0.45               & 0.17/0.30               & 0.13/0.27        & 0.21/0.38                \\ 
TUTR \cite{shi2023trajectory}      & ICCV               & 0.40/0.61               & 0.11/0.18 & 0.23/0.42               & 0.18/0.34               & 0.13/0.25        & 0.21/0.36                \\ 
PPT \cite{lin2024progressive}       & ECCV               & \uline{0.36}/\uline{0.51} & \uline{0.11}/\uline{0.15} & 0.22/0.40               & 0.17/\uline{0.30}               & 0.12/\uline{0.21}        & 0.20/\uline{0.31} \\ 
\rowcolor{green!10} AMD (ours)      & -          & \textbf{0.32}/\textbf{0.42} & \textbf{0.09}/\textbf{0.13} & \textbf{0.20}/\textbf{0.34} & \uline{0.16}/\textbf{0.26} & \uline{0.12}/\textbf{0.21} & \textbf{0.18}/\textbf{0.27} \\ 
\bottomrule
\end{tabularx}
\caption{Comparison of the performance (minADE/minFDE) of various models across all samples on the ETH/UCY dataset.}
\label{table_5}%
\end{table*}

\begin{table*}[htbp]
\centering
\footnotesize
\begin{tblr}{
  column{6-12} = {wd=1.2cm}, 
  cells = {c},
  cell{1}{1} = {r=2}{},
  cell{1}{2} = {c=4}{},
  cell{1}{6} = {c=7}{},
  hline{2} = {2-12}{},
  hline{3} = {-}{},
  rowsep = 1pt,
  row{8} = {bg=green!10},
}
\toprule
Model & Components              &         &                      &     & Performance (minADE/minFDE) &           &           &           &           &           &           \\
       & TA & Moco-DT & IC & DCL & Top 1\%     & Top 2\%   & Top 3\%   & Top 4\%   & Top 5\%   & Rest      & ALL       \\
A      & $\times$                & $\times$     & $\times$          & $\times$     & 1.55/2.41   & 1.23/1.90 & 1.06/1.65 & 0.97/1.49 & 0.90/1.37 & 0.24/0.23 & 0.28/0.29 \\
B      & $\times$                & $\checkmark$ & $\checkmark$      & $\checkmark$ & 1.47/2.05   & 1.10/1.57 & 0.95/1.35 & 0.85/1.21 & 0.79/1.10 & 0.20/0.18 & 0.23/0.22 \\
C      & $\checkmark$            & $\times$     & $\checkmark$      & $\checkmark$ & 1.30/1.79   & 0.97/1.41 & 0.85/1.22 & 0.77/1.09 & 0.71/1.00 & 0.19/0.16 & 0.22/0.20 \\
D      & $\checkmark$            & $\checkmark$ & $\times$          & $\checkmark$ & 1.38/1.95   & 1.05/1.51 & 0.90/1.30 & 0.81/1.16 & 0.75/1.06 & 0.20/0.17 & 0.23/0.21 \\
E      & $\checkmark$            & $\checkmark$ & $\checkmark$      & $\times$     & 1.45/2.02   & 1.12/1.56 & 0.95/1.32 & 0.85/1.18 & 0.77/1.08 & 0.20/0.17 & 0.23/0.21 \\
 F      & $\checkmark$            & $\checkmark$ & $\checkmark$      & $\checkmark$ & 1.08/1.66   & 0.85/1.33 & 0.75/1.15 & 0.69/1.03 & 0.64/0.95 & 0.18/0.16 & 0.21/0.21 \\
\bottomrule
\end{tblr}
\caption{Ablation results of different components on nuScenes dataset. TA means Trajectory Augmentation, IC means Iterative Clustering.}
\label{table_6}
\vspace{-6pt}
\end{table*}

\noindent \textbf{(iii) Quantitative Comparison on All Samples}.
To validate the overall effectiveness of our method, we compared it with SOTA trajectory prediction models on the nuScenes and ETH/UCY datasets. In Table~\ref{table_4}, our model outperforms others in terms of $\text{minADE}_5$ and $\text{MR}_5$ metrics, achieving improvements of 10.9\% and 5.7\%, respectively, over the leading model. Results on ETH/UCY (Table~\ref{table_5}) further confirm the AMD model’s superiority, surpassing others across all scenarios with average improvements of 5.3\% in minADE and 12.9\% in minFDE over the second-best model. These consistent and substantial performance gains clearly demonstrate our AMD model’s strong predictive capability in long-tail scenarios and robust generalization across diverse datasets and conditions.

\subsection{Ablation Studies}
Our model integrates key components that enhance its performance, evaluated through ablation studies (Table \ref{table_6}). The complete model (Model F) achieves SOTA performance across all metrics, demonstrating the strong synergistic effects among its components. Model A, with all modules removed, performs the worst, highlighting their collective importance. Model B (without trajectory augmentation) and Model E (without the decoupled contrastive learning) exhibit poor performance on top long-tail trajectories. The former struggles to capture rare trajectory patterns, while the latter increases prediction randomness due to insufficient class discrimination, indicating that trajectory augmentation and decoupled contrastive learning are crucial for long-tail learning. Other variants also show performance degradation when key components are removed, particularly on the Top 1\% samples, confirming the collaborative contribution of each component to long-tail prediction.

\begin{figure*}[t]
  \centering
\includegraphics[width=0.9\textwidth]{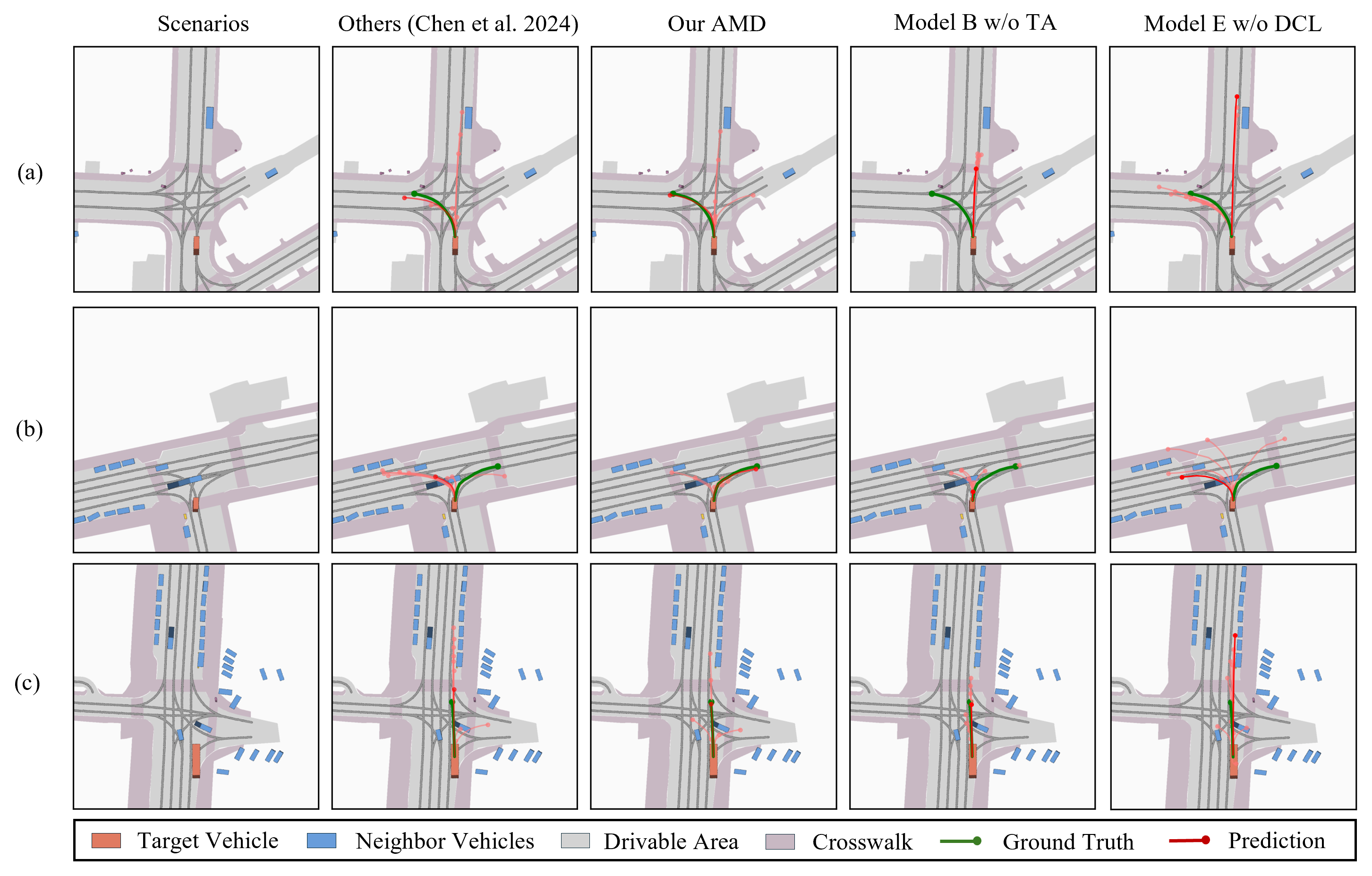} 
\vspace{-6pt}
  \caption{Qualitative results of long-tail trajectory predictions, covering various driving actions: (a) Turn left. (b) Turn right. (c) Deceleration. The red lines show the most probable trajectory, while the light red lines show the predicted multimodal trajectories. }
  \label{fig3} 
\vspace{-6pt}
\end{figure*}

\subsection{Qualitative Comparison}
Figure~\ref{fig3} presents visualization results of multimodal trajectory predictions on the nuScenes dataset under various long-tail scenarios, comparing others model \cite{chen2024q-qeanet} with ours (AMD model and its ablation variants Model B and Model E). Panels (a) and (b) depict high-curvature vehicle turning trajectories, while Panel (c) shows a trajectory with distinct deceleration actions. The results demonstrate that AMD accurately predicts these complex trajectories and generates additional plausible options. Compared to Model B and Model E, the trajectory augmentation (TA) strategy enhances generalization to complex dynamics by producing diverse samples, effectively capturing geometric features of maneuvers like turns. Meanwhile, decoupled contrastive learning (DCL) improves differentiation of rare trajectories by separating positive and negative sample representations, reducing prediction randomness. This mechanism enables AMD to maintain accuracy and model multimodal uncertainty effectively in long-tail distributions.

\subsection{Inference Time Comparison}
To demonstrate the efficiency of our AMD model, we conducted a comparative experiment on inference times using the nuScenes dataset, with VisionTrap tested on an RTX 3090 Ti GPU and the other models, including ours, evaluated on an RTX 3090 GPU. As shown in Table~\ref{tab7}, AMD exhibits a clear advantage in inference speed. The results highlight that our model significantly reduces inference time while maintaining accuracy, making it well-suited for real-time autonomous driving.

\begin{table}[htbp]
    \centering
    \footnotesize
    \begin{tblr}{
    cells = {c},
    hline{1,2} = {-}{},
    hline{6,7} = {-}{},
    row{6} = {bg=green!10},
    rowsep = 1pt,
    }
        Model & Inference Time (ms) \\ 
        Trajectron++ \cite{salzmann2020trajectron++}  & 38 \\
        PGP \cite{deo2022multimodal} & 215 \\
        LAformer \cite{liu2024laformer} & 115 \\
        VisionTrap \cite{moon2024visiontrap} & 53 \\
        \textbf{AMD} & 14 \\
    \end{tblr}
\caption{Inference time comparison on nuScenes dataset.}
\label{tab7}%
\end{table}
\vspace{-6pt}

\subsection{Feature Space Visualisation}
\begin{figure}[htbp]
  \centering
  \includegraphics[width=1\linewidth]{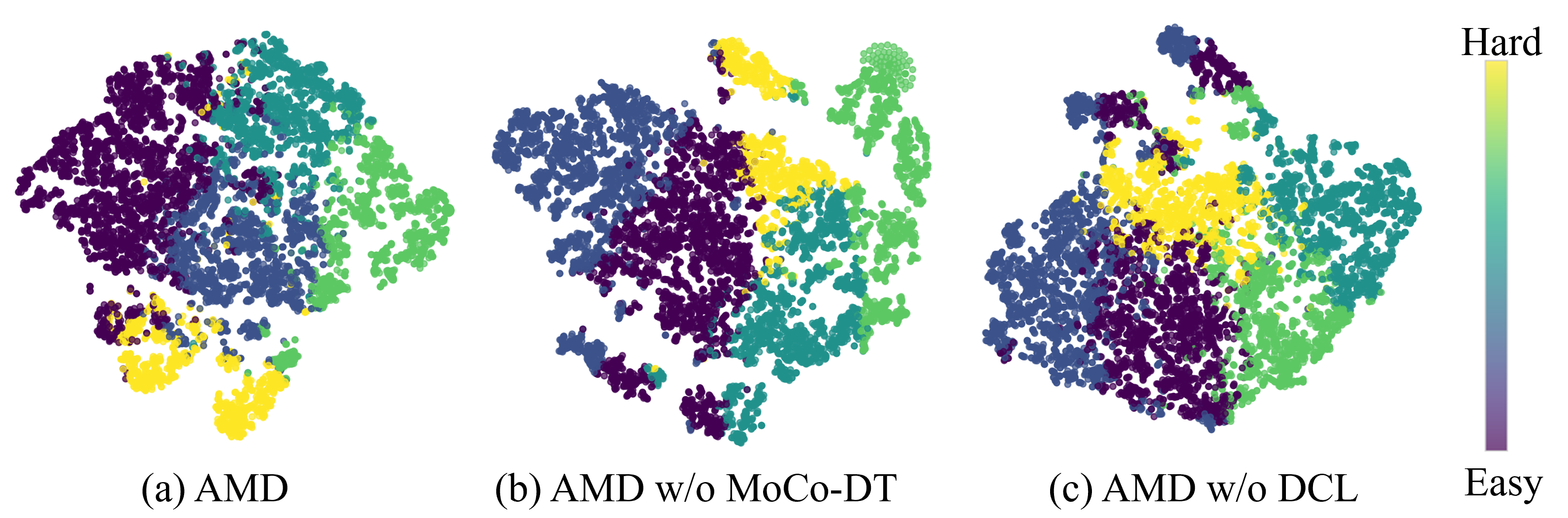}
   \caption{Visualization of feature spaces for different variants.}
   \label{fig:vis1}
\end{figure}

We conducted feature space visualization by applying t-SNE on nuScenes dataset to reduce the extracted features into a two-dimensional space for analysis. As shown in Figure~\ref{fig:vis1}a, our model enhances cluster compactness and tail separation, clearly separating head and tail patterns while forming distinct clusters for tail patterns. In contrast, removing MoCo-DT (Figure~\ref{fig:vis1}b) disperses hard samples, cluttering tail patterns, while removing DCL (Figure~\ref{fig:vis1}c) increases head-tail overlap. This confirms MoCo-DT boosts tail representation and DCL mitigates head-class dominance via balanced learning.

\vspace{-6pt}

\subsection{Hyperparameter Sensitivity Study}

We conducted a sensitivity analysis of the hyperparameters $\lambda_1$, $\lambda_2$, $m_e$, $m_m$, and $m_l$ on the nuScenes dataset, with results presented in Table~\ref{tab8}. The parameters $\lambda_1$ and $\lambda_2$ influence the model’s ability to address long-tail trajectories by balancing loss contributions, while $m_e$, $m_m$, and $m_l$ primarily affect model stability. The optimal combination yielding the best performance was selected.

\begin{table}[htbp]

    \centering
    \footnotesize
    \begin{tblr}{
    column{1} = {wd=0.45cm},
    column{2} = {wd=1.9cm},
    column{3-5} = {wd=0.89cm},
    column{6} = {wd=0.85cm},
    cells = {c},
    hline{1,2,6,9} = {-}{},
    rowsep = 1.2pt,
    }
       \SetCell[c=2]{c}{Hyperparameter Setting} & & Top 1\% & Top 2\% & Top 3\% & ALL \\
        \SetCell[r=4]{c} {\(\begin{array}{c} \lambda_1,  \lambda_2 \end{array}\)}
        & 1.0, 0.5   & 1.10/1.77 & 0.87/1.39 & 0.77/1.19 & 0.22/0.22 \\
        & 1.0, 0.1   & \textbf{1.08}/\textbf{1.66} & \textbf{0.85}/\textbf{1.33} & \textbf{0.75}/\textbf{1.15} & \textbf{0.21}/\textbf{0.21} \\
        & 0.5, 0.5 & 1.11/1.68 & 0.87/1.36 & 0.77/1.15 & 0.21/0.22 \\
        & 0.5, 0.1 & 1.26/2.00 & 0.97/1.53 & 0.87/1.33 & 0.22/0.24 \\
        \SetCell[r=3]{c} {\(\begin{array}{c} m_e, \\ m_m, \\ m_l \end{array}\)}
        & 0.90, 0.95, 0.999  & 1.16/2.04 & 0.94/1.60 & 0.83/1.37 & 0.25/0.26 \\
        & 0.90, 0.99, 0.999  & 1.14/1.79 & 0.91/1.38 & 0.77/1.16 & 0.22/0.22 \\
        & 0.95, 0.99, 0.999  & \textbf{1.08}/\textbf{1.66} & \textbf{0.85}/\textbf{1.33} & \textbf{0.75}/\textbf{1.15} & \textbf{0.21}/\textbf{0.21} \\
    \end{tblr}
    \caption{Hyperparameter sensitivity analysis on nuScenes dataset.}
    \label{tab8}
\vspace{-12pt}
\end{table}



%% file: sec/5_conclusion.tex
\section{Conclusion}
\label{sec:Conclusion}

In this paper, we propose an Adaptive Momentum and Decoupled Contrastive Learning framework (AMD) tailored for robust trajectory prediction in challenging long-tail scenarios. Leveraging a novel combination of unsupervised and supervised contrastive learning, AMD effectively enhances predictive performance on rare trajectory patterns while maintaining high accuracy on general trajectory distributions. Additionally, our random trajectory augmentation and online iterative learning strategies significantly boost the model's adaptability, allowing it to robustly handle complex and diverse spatiotemporal dynamics. Experimental results demonstrate that AMD consistently surpasses SOTA methods on long-tail subsets and achieves competitive overall accuracy across multiple datasets.

%% file: main.bbl
\begin{thebibliography}{62}
\providecommand{\natexlab}[1]{#1}
\providecommand{\url}[1]{\texttt{#1}}
\expandafter\ifx\csname urlstyle\endcsname\relax
  \providecommand{\doi}[1]{doi: #1}\else
  \providecommand{\doi}{doi: \begingroup \urlstyle{rm}\Url}\fi

\bibitem[Alahi et~al.(2016)Alahi, Goel, Ramanathan, Robicquet, Fei-Fei, and Savarese]{alahi2016social-lstm}
Alexandre Alahi, Kratarth Goel, Vignesh Ramanathan, Alexandre Robicquet, Li Fei-Fei, and Silvio Savarese.
\newblock Social lstm: Human trajectory prediction in crowded spaces.
\newblock In \emph{CVPR}, pages 961--971, 2016.

\bibitem[Bae et~al.(2022)Bae, Park, and Jeon]{bae2022non}
Inhwan Bae, Jin-Hwi Park, and Hae-Gon Jeon.
\newblock Non-probability sampling network for stochastic human trajectory prediction.
\newblock In \emph{CVPR}, pages 6477--6487, 2022.

\bibitem[Caesar et~al.(2020)Caesar, Bankiti, Lang, Vora, Liong, Xu, Krishnan, Pan, Baldan, and Beijbom]{caesar2020nuscenes}
Holger Caesar, Varun Bankiti, Alex~H Lang, Sourabh Vora, Venice~Erin Liong, Qiang Xu, Anush Krishnan, Yu Pan, Giancarlo Baldan, and Oscar Beijbom.
\newblock nuscenes: A multimodal dataset for autonomous driving.
\newblock In \emph{CVPR}, pages 11621--11631, 2020.

\bibitem[Cao et~al.(2019)Cao, Wei, Gaidon, Arechiga, and Ma]{cao2019learning}
Kaidi Cao, Colin Wei, Adrien Gaidon, Nikos Arechiga, and Tengyu Ma.
\newblock Learning imbalanced datasets with label-distribution-aware margin loss.
\newblock \emph{Advances in neural information processing systems}, 32, 2019.

\bibitem[Chang et~al.(2023)Chang, Qi, Liang, and Tanin]{chang2023contrastive}
Yanchuan Chang, Jianzhong Qi, Yuxuan Liang, and Egemen Tanin.
\newblock Contrastive trajectory similarity learning with dual-feature attention.
\newblock In \emph{2023 IEEE 39th International conference on data engineering (ICDE)}, pages 2933--2945. IEEE, 2023.

\bibitem[Chen et~al.(2024)Chen, Wang, Wang, and Cai]{chen2024q-qeanet}
Jiuyu Chen, Zhongli Wang, Jian Wang, and Baigen Cai.
\newblock Q-eanet: Implicit social modeling for trajectory prediction via experience-anchored queries.
\newblock \emph{IET Intelligent Transport Systems}, 18\penalty0 (6):\penalty0 1004--1015, 2024.

\bibitem[Chen et~al.(2020)Chen, Kornblith, Norouzi, and Hinton]{chen2020simple-clr}
Ting Chen, Simon Kornblith, Mohammad Norouzi, and Geoffrey Hinton.
\newblock A simple framework for contrastive learning of visual representations.
\newblock In \emph{International conference on machine learning}, pages 1597--1607. PMLR, 2020.

\bibitem[Cui et~al.(2019)Cui, Jia, Lin, Song, and Belongie]{cui2019class}
Yin Cui, Menglin Jia, Tsung-Yi Lin, Yang Song, and Serge Belongie.
\newblock Class-balanced loss based on effective number of samples.
\newblock In \emph{CVPR}, pages 9268--9277, 2019.

\bibitem[Deo and Trivedi(2020)]{deo2020trajectory-P2T}
Nachiket Deo and Mohan~M Trivedi.
\newblock Trajectory forecasts in unknown environments conditioned on grid-based plans.
\newblock \emph{arXiv preprint arXiv:2001.00735}, 2020.

\bibitem[Deo et~al.(2022)Deo, Wolff, and Beijbom]{deo2022multimodal}
Nachiket Deo, Eric Wolff, and Oscar Beijbom.
\newblock Multimodal trajectory prediction conditioned on lane-graph traversals.
\newblock In \emph{Conference on Robot Learning}, pages 203--212. PMLR, 2022.

\bibitem[Gao et~al.(2020)Gao, Sun, Zhao, Shen, Anguelov, Li, and Schmid]{gao2020vectornet}
Jiyang Gao, Chen Sun, Hang Zhao, Yi Shen, Dragomir Anguelov, Congcong Li, and Cordelia Schmid.
\newblock Vectornet: Encoding hd maps and agent dynamics from vectorized representation.
\newblock In \emph{CVPR}, pages 11525--11533, 2020.

\bibitem[Gilles et~al.(2022)Gilles, Sabatini, Tsishkou, Stanciulescu, and Moutarde]{gilles2022gohome}
Thomas Gilles, Stefano Sabatini, Dzmitry Tsishkou, Bogdan Stanciulescu, and Fabien Moutarde.
\newblock Gohome: Graph-oriented heatmap output for future motion estimation.
\newblock In \emph{2022 international conference on robotics and automation (ICRA)}, pages 9107--9114. IEEE, 2022.

\bibitem[Gu et~al.(2022)Gu, Chen, Li, Lin, Rao, Zhou, and Lu]{gu2022stochastic}
Tianpei Gu, Guangyi Chen, Junlong Li, Chunze Lin, Yongming Rao, Jie Zhou, and Jiwen Lu.
\newblock Stochastic trajectory prediction via motion indeterminacy diffusion.
\newblock In \emph{CVPR}, pages 17113--17122, 2022.

\bibitem[Han et~al.(2005)Han, Wang, and Mao]{han2005borderline-resample}
Hui Han, Wen-Yuan Wang, and Bing-Huan Mao.
\newblock Borderline-smote: a new over-sampling method in imbalanced data sets learning.
\newblock In \emph{International conference on intelligent computing}, pages 878--887. Springer, 2005.

\bibitem[Hartigan and Wong(1979)]{hartigan1979algorithm-kmeans}
John~A Hartigan and Manchek~A Wong.
\newblock Algorithm as 136: A k-means clustering algorithm.
\newblock \emph{Journal of the royal statistical society. series c (applied statistics)}, 28\penalty0 (1):\penalty0 100--108, 1979.

\bibitem[He et~al.(2020)He, Fan, Wu, Xie, and Girshick]{he2020momentum-moco}
Kaiming He, Haoqi Fan, Yuxin Wu, Saining Xie, and Ross Girshick.
\newblock Momentum contrast for unsupervised visual representation learning.
\newblock In \emph{CVPR}, pages 9729--9738, 2020.

\bibitem[Huang et~al.(2021)Huang, Zhu, Xiao, and Liu]{huang2021bayonet-GRU}
Mengyang Huang, Menggang Zhu, Yunpeng Xiao, and Yanbing Liu.
\newblock Bayonet-corpus: a trajectory prediction method based on bayonet context and bidirectional gru.
\newblock \emph{Digital Communications and Networks}, 7\penalty0 (1):\penalty0 72--81, 2021.

\bibitem[Khosla et~al.(2020)Khosla, Teterwak, Wang, Sarna, Tian, Isola, Maschinot, Liu, and Krishnan]{khosla2020supervised-cl}
Prannay Khosla, Piotr Teterwak, Chen Wang, Aaron Sarna, Yonglong Tian, Phillip Isola, Aaron Maschinot, Ce Liu, and Dilip Krishnan.
\newblock Supervised contrastive learning.
\newblock \emph{Advances in neural information processing systems}, 33:\penalty0 18661--18673, 2020.

\bibitem[Kim et~al.(2021)Kim, Park, Lee, Khoshimjonov, Kum, Kim, Kim, and Choi]{kim2021lapred}
ByeoungDo Kim, Seong~Hyeon Park, Seokhwan Lee, Elbek Khoshimjonov, Dongsuk Kum, Junsoo Kim, Jeong~Soo Kim, and Jun~Won Choi.
\newblock Lapred: Lane-aware prediction of multi-modal future trajectories of dynamic agents.
\newblock In \emph{CVPR}, pages 14636--14645, 2021.

\bibitem[Lan et~al.(2024)Lan, Ren, Yu, Liu, Li, Wang, and Cui]{lan2024hi-scl}
Zhengxing Lan, Yilong Ren, Haiyang Yu, Lingshan Liu, Zhenning Li, Yinhai Wang, and Zhiyong Cui.
\newblock Hi-scl: Fighting long-tailed challenges in trajectory prediction with hierarchical wave-semantic contrastive learning.
\newblock \emph{Transportation Research Part C: Emerging Technologies}, 165:\penalty0 104735, 2024.

\bibitem[Leal-Taix{\'e} et~al.(2014)Leal-Taix{\'e}, Fenzi, Kuznetsova, Rosenhahn, and Savarese]{leal2014learning}
Laura Leal-Taix{\'e}, Michele Fenzi, Alina Kuznetsova, Bodo Rosenhahn, and Silvio Savarese.
\newblock Learning an image-based motion context for multiple people tracking.
\newblock In \emph{CVPR}, pages 3542--3549, 2014.

\bibitem[Li et~al.(2022)Li, Han, Li, Fu, and Zhang]{li2022longtaildis}
Bolian Li, Zongbo Han, Haining Li, Huazhu Fu, and Changqing Zhang.
\newblock Trustworthy long-tailed classification.
\newblock In \emph{CVPR}, pages 6970--6979, 2022.

\bibitem[Li et~al.(2023)Li, Liu, Li, Yu, Cao, Qiu, Hu, Wang, and Jiao]{li2023graph-risk}
Xueke Li, Jiaxin Liu, Jun Li, Wenhao Yu, Zhong Cao, Shaobo Qiu, Jia Hu, Hong Wang, and Xiaohong Jiao.
\newblock Graph structure-based implicit risk reasoning for long-tail scenarios of automated driving.
\newblock In \emph{2023 4th International Conference on Big Data, Artificial Intelligence and Internet of Things Engineering (ICBAIE)}, pages 415--420. IEEE, 2023.

\bibitem[Liao et~al.(2024{\natexlab{a}})Liao, Li, Li, Kong, Wang, Wang, Guan, Tam, and Li]{liao2024cdstraj}
Haicheng Liao, Xuelin Li, Yongkang Li, Hanlin Kong, Chengyue Wang, Bonan Wang, Yanchen Guan, K Tam, and Zhenning Li.
\newblock Cdstraj: Characterized diffusion and spatial-temporal interaction network for trajectory prediction in autonomous driving.
\newblock In \emph{IJCAI}, pages 7331--7339, 2024{\natexlab{a}}.

\bibitem[Liao et~al.(2024{\natexlab{b}})Liao, Li, Shen, Zeng, Liao, Li, and Xu]{liao2024bat-trans}
Haicheng Liao, Zhenning Li, Huanming Shen, Wenxuan Zeng, Dongping Liao, Guofa Li, and Chengzhong Xu.
\newblock Bat: Behavior-aware human-like trajectory prediction for autonomous driving.
\newblock In \emph{AAAI}, pages 10332--10340, 2024{\natexlab{b}}.

\bibitem[Liao et~al.(2024{\natexlab{c}})Liao, Li, Wang, Shen, Liao, Wang, Li, and Xu]{liao2024mftraj}
Haicheng Liao, Zhenning Li, Chengyue Wang, Huanming Shen, Dongping Liao, Bonan Wang, Guofa Li, and Chengzhong Xu.
\newblock Mftraj: Map-free, behavior-driven trajectory prediction for autonomous driving.
\newblock In \emph{IJCAI}, 2024{\natexlab{c}}.

\bibitem[Liao et~al.(2024{\natexlab{d}})Liao, Li, Wang, Wang, Kong, Guan, Li, and Cui]{liao2024cognitive}
Haicheng Liao, Zhenning Li, Chengyue Wang, Bonan Wang, Hanlin Kong, Yanchen Guan, Guofa Li, and Zhiyong Cui.
\newblock A cognitive-driven trajectory prediction model for autonomous driving in mixed autonomy environments.
\newblock In \emph{IJCAI}, 2024{\natexlab{d}}.

\bibitem[Liao et~al.(2024{\natexlab{e}})Liao, Wang, Li, Li, Wang, Li, and Xu]{liao2024physics}
Haicheng Liao, Chengyue Wang, Zhenning Li, Yongkang Li, Bonan Wang, Guofa Li, and Chengzhong Xu.
\newblock Physics-informed trajectory prediction for autonomous driving under missing observation.
\newblock In \emph{IJCAI}, 2024{\natexlab{e}}.

\bibitem[Liao et~al.(2025{\natexlab{a}})Liao, Kong, Wang, Wang, Ye, He, Xu, and Li]{liao2025cot}
Haicheng Liao, Hanlin Kong, Bonan Wang, Chengyue Wang, Wang Ye, Zhengbing He, Chengzhong Xu, and Zhenning Li.
\newblock Cot-drive: Efficient motion forecasting for autonomous driving with llms and chain-of-thought prompting.
\newblock \emph{IEEE Transactions on Artificial Intelligence}, 2025{\natexlab{a}}.

\bibitem[Liao et~al.(2025{\natexlab{b}})Liao, Li, Zhang, Li, and Xu]{liao2025toward}
Haicheng Liao, Zhenning Li, Guohui Zhang, Keqiang Li, and Chengzhong Xu.
\newblock Toward human-like trajectory prediction for autonomous driving: A behavior-centric approach.
\newblock \emph{Transportation Science}, 2025{\natexlab{b}}.

\bibitem[Liao et~al.(2025{\natexlab{c}})Liao, Wang, Zhu, Ren, Gao, Li, Xu, and Li]{liao2025minds}
Haicheng Liao, Chengyue Wang, Kaiqun Zhu, Yilong Ren, Bolin Gao, Shengbo~Eben Li, Chengzhong Xu, and Zhenning Li.
\newblock Minds on the move: Decoding trajectory prediction in autonomous driving with cognitive insights.
\newblock \emph{IEEE Transactions on Intelligent Transportation Systems}, 2025{\natexlab{c}}.

\bibitem[Lin et~al.(2000)Lin, Ulsoy, and LeBlanc]{lin2000vehicle-yuce-1}
Chiu-Feng Lin, A~Galip Ulsoy, and David~J LeBlanc.
\newblock Vehicle dynamics and external disturbance estimation for vehicle path prediction.
\newblock \emph{IEEE Transactions on Control Systems Technology}, 8\penalty0 (3):\penalty0 508--518, 2000.

\bibitem[Lin et~al.(2024)Lin, Liang, Lai, and Hu]{lin2024progressive}
Xiaotong Lin, Tianming Liang, Jianhuang Lai, and Jian-Fang Hu.
\newblock Progressive pretext task learning for human trajectory prediction.
\newblock In \emph{ECCV}, pages 197--214. Springer, 2024.

\bibitem[Liu et~al.(2024)Liu, Cheng, Chen, Broszio, Li, Zhao, Sester, and Yang]{liu2024laformer}
Mengmeng Liu, Hao Cheng, Lin Chen, Hellward Broszio, Jiangtao Li, Runjiang Zhao, Monika Sester, and Michael~Ying Yang.
\newblock Laformer: Trajectory prediction for autonomous driving with lane-aware scene constraints.
\newblock In \emph{CVPR}, pages 2039--2049, 2024.

\bibitem[Liu et~al.(2019)Liu, Miao, Zhan, Wang, Gong, and Yu]{longtail-2}
Ziwei Liu, Zhongqi Miao, Xiaohang Zhan, Jiayun Wang, Boqing Gong, and Stella~X Yu.
\newblock Large-scale long-tailed recognition in an open world.
\newblock In \emph{CVPR}, pages 2537--2546, 2019.

\bibitem[Liu et~al.(2022)Liu, Miao, Zhan, Wang, Gong, and Stella]{longtail-1}
Ziwei Liu, Zhongqi Miao, Xiaohang Zhan, Jiayun Wang, Boqing Gong, and X~Yu Stella.
\newblock Open long-tailed recognition in a dynamic world.
\newblock \emph{IEEE TPAMI}, 46\penalty0 (3):\penalty0 1836--1851, 2022.

\bibitem[Makansi et~al.(2021)Makansi, Cicek, Marrakchi, and Brox]{makansi2021on-exposing}
Osama Makansi, {\"O}zg{\"u}n Cicek, Yassine Marrakchi, and Thomas Brox.
\newblock On exposing the challenging long tail in future prediction of traffic actors.
\newblock In \emph{ICCV}, pages 13147--13157, 2021.

\bibitem[Mangalam et~al.(2020)Mangalam, Girase, Agarwal, Lee, Adeli, Malik, and Gaidon]{mangalam2020not}
Karttikeya Mangalam, Harshayu Girase, Shreyas Agarwal, Kuan-Hui Lee, Ehsan Adeli, Jitendra Malik, and Adrien Gaidon.
\newblock It is not the journey but the destination: Endpoint conditioned trajectory prediction.
\newblock In \emph{ECCV}, pages 759--776. Springer, 2020.

\bibitem[Mohamed et~al.(2020)Mohamed, Qian, Elhoseiny, and Claudel]{mohamed2020social}
Abduallah Mohamed, Kun Qian, Mohamed Elhoseiny, and Christian Claudel.
\newblock Social-stgcnn: A social spatio-temporal graph convolutional neural network for human trajectory prediction.
\newblock In \emph{CVPR}, pages 14424--14432, 2020.

\bibitem[Moon et~al.(2024)Moon, Woo, Park, Jung, Mahjourian, Chi, Lim, Kim, and Kim]{moon2024visiontrap}
Seokha Moon, Hyun Woo, Hongbeen Park, Haeji Jung, Reza Mahjourian, Hyung-gun Chi, Hyerin Lim, Sangpil Kim, and Jinkyu Kim.
\newblock Visiontrap: Vision-augmented trajectory prediction guided by textual descriptions.
\newblock In \emph{ECCV}, pages 361--379. Springer, 2024.

\bibitem[Pellegrini et~al.(2009)Pellegrini, Ess, Schindler, and Van~Gool]{pellegrini2009you}
Stefano Pellegrini, Andreas Ess, Konrad Schindler, and Luc Van~Gool.
\newblock You'll never walk alone: Modeling social behavior for multi-target tracking.
\newblock In \emph{ICCV}, pages 261--268. IEEE, 2009.

\bibitem[Reed(2001)]{reed2001pareto}
William~J Reed.
\newblock The pareto, zipf and other power laws.
\newblock \emph{Economics letters}, 74\penalty0 (1):\penalty0 15--19, 2001.

\bibitem[Ross and Doll{\'a}r(2017)]{ross2017focal-loss}
T-YLPG Ross and GKHP Doll{\'a}r.
\newblock Focal loss for dense object detection.
\newblock In \emph{CVPR}, pages 2980--2988, 2017.

\bibitem[Salzmann et~al.(2020)Salzmann, Ivanovic, Chakravarty, and Pavone]{salzmann2020trajectron++}
Tim Salzmann, Boris Ivanovic, Punarjay Chakravarty, and Marco Pavone.
\newblock Trajectron++: Dynamically-feasible trajectory forecasting with heterogeneous data.
\newblock In \emph{ECCV}, pages 683--700. Springer, 2020.

\bibitem[Shen et~al.(2016)Shen, Lin, and Huang]{shen2016relay}
Li Shen, Zhouchen Lin, and Qingming Huang.
\newblock Relay backpropagation for effective learning of deep convolutional neural networks.
\newblock In \emph{ECCV}, pages 467--482. Springer, 2016.

\bibitem[Shi et~al.(2023)Shi, Wang, Zhou, and Hua]{shi2023trajectory}
Liushuai Shi, Le Wang, Sanping Zhou, and Gang Hua.
\newblock Trajectory unified transformer for pedestrian trajectory prediction.
\newblock In \emph{ICCV}, pages 9675--9684, 2023.

\bibitem[Thuremella et~al.(2024)Thuremella, Ince, and Kunze]{thuremella2024risk}
Divya Thuremella, Lewis Ince, and Lars Kunze.
\newblock Risk-aware trajectory prediction by incorporating spatio-temporal traffic interaction analysis.
\newblock In \emph{2024 IEEE International Conference on Robotics and Automation (ICRA)}, pages 14421--14427. IEEE, 2024.

\bibitem[Wang et~al.(2025{\natexlab{a}})Wang, Liao, Wang, Rao, Guan, Yu, Zhang, Lai, Xu, and Li]{wang2025beyond}
Bonan Wang, Haicheng Liao, Chengyue Wang, Bin Rao, Yanchen Guan, Guyang Yu, Jiaxun Zhang, Songning Lai, Chengzhong Xu, and Zhenning Li.
\newblock Beyond patterns: Harnessing causal logic for autonomous driving trajectory prediction.
\newblock \emph{arXiv preprint arXiv:2505.06856}, 2025{\natexlab{a}}.

\bibitem[Wang et~al.(2025{\natexlab{b}})Wang, Liao, Li, and Xu]{wang2025wake}
Chengyue Wang, Haicheng Liao, Zhenning Li, and Chengzhong Xu.
\newblock Wake: Towards robust and physically feasible trajectory prediction for autonomous vehicles with wavelet and kinematics synergy.
\newblock \emph{PAMI}, 2025{\natexlab{b}}.

\bibitem[Wang et~al.(2025{\natexlab{c}})Wang, Liao, Wang, Guan, Rao, Pu, Cui, Xu, and Li]{wang2025nest}
Chengyue Wang, Haicheng Liao, Bonan Wang, Yanchen Guan, Bin Rao, Ziyuan Pu, Zhiyong Cui, Cheng-Zhong Xu, and Zhenning Li.
\newblock Nest: A neuromodulated small-world hypergraph trajectory prediction model for autonomous driving.
\newblock In \emph{AAAI}, pages 808--816, 2025{\natexlab{c}}.

\bibitem[Wang et~al.(2025{\natexlab{d}})Wang, Liao, Zhu, Zhang, and Li]{wang2025dynamics}
Chengyue Wang, Haicheng Liao, Kaiqun Zhu, Guohui Zhang, and Zhenning Li.
\newblock A dynamics-enhanced learning model for multi-horizon trajectory prediction in autonomous vehicles.
\newblock \emph{Information Fusion}, 118:\penalty0 102924, 2025{\natexlab{d}}.

\bibitem[Wang et~al.(2024)Wang, Hu, Song, and Li]{WOS:001270539300037}
Ruiping Wang, Zhijian Hu, Xiao Song, and Wenxin Li.
\newblock Trajectory distribution aware graph convolutional network for trajectory prediction considering spatio-temporal interactions and scene information.
\newblock \emph{IEEE TRANSACTIONS ON KNOWLEDGE AND DATA ENGINEERING}, 36\penalty0 (8):\penalty0 4304--4316, 2024.

\bibitem[Wang et~al.(2023)Wang, Zhang, Bai, and Xue]{wang2023fend}
Yuning Wang, Pu Zhang, Lei Bai, and Jianru Xue.
\newblock Fend: A future enhanced distribution-aware contrastive learning framework for long-tail trajectory prediction.
\newblock In \emph{CVPR}, pages 1400--1409, 2023.

\bibitem[Wong et~al.(2022)Wong, Xia, Hong, Peng, Yuan, Cao, Yang, and You]{wong2022view-yuce-2}
Conghao Wong, Beihao Xia, Ziming Hong, Qinmu Peng, Wei Yuan, Qiong Cao, Yibo Yang, and Xinge You.
\newblock View vertically: A hierarchical network for trajectory prediction via fourier spectrums.
\newblock In \emph{ECCV}, pages 682--700. Springer, 2022.

\bibitem[Xu et~al.(2023)Xu, Hayet, and Karamouzas]{xu2023context-VAE}
Pei Xu, Jean-Bernard Hayet, and Ioannis Karamouzas.
\newblock Context-aware timewise vaes for real-time vehicle trajectory prediction.
\newblock \emph{IEEE Robotics and Automation Letters}, 2023.

\bibitem[Xu and Fu(2024)]{xu2024adapting-AFormer}
Yi Xu and Yun Fu.
\newblock Adapting to length shift: Flexilength network for trajectory prediction.
\newblock In \emph{CVPR}, pages 15226--15237, 2024.

\bibitem[Xu et~al.(2022)Xu, Wang, Wang, and Fu]{xu2022adaptive-GNN}
Yi Xu, Lichen Wang, Yizhou Wang, and Yun Fu.
\newblock Adaptive trajectory prediction via transferable gnn.
\newblock In \emph{CVPR}, pages 6520--6531, 2022.

\bibitem[Xuan and Zhang(2024)]{xuan2024decoupled}
Shiyu Xuan and Shiliang Zhang.
\newblock Decoupled contrastive learning for long-tailed recognition.
\newblock In \emph{AAAI}, pages 6396--6403, 2024.

\bibitem[Yuan et~al.(2021)Yuan, Weng, Ou, and Kitani]{yuan2021agentformer}
Ye Yuan, Xinshuo Weng, Yanglan Ou, and Kris~M Kitani.
\newblock Agentformer: Agent-aware transformers for socio-temporal multi-agent forecasting.
\newblock In \emph{ICCV}, pages 9813--9823, 2021.

\bibitem[Zhang et~al.(2024)Zhang, Pourkeshavarz, and Rasouli]{zhang2024tract}
Junrui Zhang, Mozhgan Pourkeshavarz, and Amir Rasouli.
\newblock Tract: A training dynamics aware contrastive learning framework for long-tail trajectory prediction.
\newblock \emph{arXiv preprint arXiv:2404.12538}, 2024.

\bibitem[Zhang et~al.(2025)Zhang, Yang, Li, Andersson, and Jensfelt]{zhang2025seflow}
Qingwen Zhang, Yi Yang, Peizheng Li, Olov Andersson, and Patric Jensfelt.
\newblock Seflow: A self-supervised scene flow method in autonomous driving.
\newblock In \emph{ECCV}, pages 353--369. Springer, 2025.

\bibitem[Zhou et~al.(2022)Zhou, Cao, Xu, Deng, Liu, Jiang, and Yang]{zhou2022long}
Weitao Zhou, Zhong Cao, Yunkang Xu, Nanshan Deng, Xiaoyu Liu, Kun Jiang, and Diange Yang.
\newblock Long-tail prediction uncertainty aware trajectory planning for self-driving vehicles.
\newblock In \emph{2022 IEEE 25th International Conference on Intelligent Transportation Systems (ITSC)}, pages 1275--1282. IEEE, 2022.

\end{thebibliography}
